%% file: RAL2017_peduncle_detection.tex
\documentclass[letterpaper, 10 pt, journal, twoside]{IEEEtran} 
\IEEEoverridecommandlockouts                              

\usepackage{graphicx} 
\usepackage{times} 
\usepackage{amsmath} 
\usepackage{amssymb}  
\usepackage{verbatim} 
\usepackage{cite}
\usepackage{subfigure}
\usepackage[hyphens]{url}
\usepackage[ruled,vlined]{algorithm2e}
\usepackage{float}
\usepackage{tabulary}
\usepackage{tabularx}
\usepackage{lipsum,amsmath,multicol}
\usepackage{tensor}
\usepackage{multirow}
\usepackage{flushend}

\input{./extern/pic-common}

\usepackage{color}

\def\NOTE#1{{\bf [NOTE:} {\it\color{blue}{#1}}{\bf ]}.}

\definecolor{CommentRed}{rgb}{0.7,0,0}
\definecolor{CommentBlue}{rgb}{0,0,0.7}
\definecolor{CommentDG}{rgb}{0,0.6,0}

\newcommand{\footnoteref}[1]{\textsuperscript{\ref{#1}}}

\makeatletter
\newenvironment{myalign*}{%
  \setlength{\mathindent}{0pt}%
  \setlength{\abovedisplayskip}{-\baselineskip}%
  \setlength{\abovedisplayshortskip}{\abovedisplayskip}%
  \start@align\@ne\st@rredtrue\m@ne
}%
{\endalign}
\makeatother

\title{Peduncle Detection of Sweet Pepper for Autonomous Crop Harvesting - Combined Colour and 3D Information} 

\author{Inkyu Sa$^{1*}$, Chris Lehnert$^{2}$, Andrew English$^{2}$, Chris McCool$^{2}$, Feras Dayoub$^{2}$, Ben Upcroft$^{2}$, Tristan Perez$^{2}$%
\thanks{*This project has received funding from the Queensland Department of Agriculture and Fisheries and QUT Strategic Investment in Farm Robotics Program.}
\thanks{$^{1*}$Autonomous Systems Lab., Department of Mechanical and Process Engineering, ETH Zurich, Zurich, Switzerland. ${}^*$This work was done when Inkyu Sa was in QUT.
        {\tt\small inkyu.sa@mavt.ethz.ch}}%
\thanks{$^{2} $Science and Engineering Faculty, Queensland University of Technology, Brisbane, Australia.
        {\tt\small c.lehnert, a.english, c.mccool, feras.dayoub, ben.upcroft, tristan.perez@qut.edu.au}}%
}
\begin{document}
\maketitle

\begin{abstract}
This paper presents a 3D visual detection method for the challenging task of detecting peduncles of sweet peppers (Capsicum annuum) in the field. Cutting the peduncle cleanly is one of the most difficult stages of the harvesting process, where the peduncle is the part of the crop that attaches it to the main stem of the plant. Accurate peduncle detection in 3D space is therefore a vital step in reliable autonomous harvesting of sweet peppers, as this can lead to precise cutting while avoiding damage to the surrounding plant. This paper makes use of both colour and geometry information acquired from an RGB-D sensor and utilises a supervised-learning approach for the peduncle detection task. The performance of the proposed method is demonstrated and evaluated using qualitative and quantitative results (the Area-Under-the-Curve (AUC) of the detection precision-recall curve). We are able to achieve an AUC of 0.71 for peduncle detection on field-grown sweet peppers. 
We release a set of manually annotated 3D sweet pepper and peduncle images to assist the research community in performing further research on this topic.
\end{abstract}

\section{INTRODUCTION}
\label{sec:intro}
\IEEEPARstart{A}{utomating} agricultural processes such as planting, harvesting, weeding and inspection will play a key role in helping to improve farm productivity, increase crop quality and reduce input costs \cite{kondo2011agricultural}. In particular, the harvesting of high value crops within horticulture still demands a large amount of hand labour costs due to the dexterity required to achieve the task. For these reasons, automating the task of harvesting is of great interest to the horticultural and robotics industries.
Recent research in robotics has made impressive progress towards the goal of developing viable broad-acre~\cite{english2014vision,Ball:2016aa} and horticultural robots~\cite{sa2015visual,mccool2016visual,Lehnert:2016aa,bac2014harvesting,Kitamura:2005aa}.

\begin{figure}[!t]
\begin{center}
\includegraphics[width=\columnwidth]{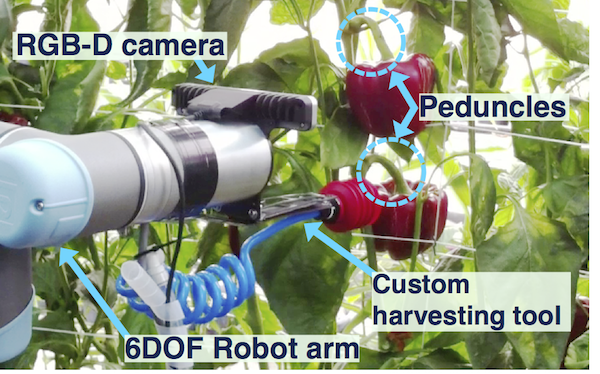}
\end{center}
\vspace{-10pt}
	\caption{Sweet pepper picking in operation showing a robotics arm equipped with an end-effector tool to harvest the pepper by cutting its peduncle. The photo highlights the presence of occlusions and varying lighting conditions of peduncles and sweet peppers grown in a field environment.}
	\label{fig:first_image}
\vspace{-15pt}
\end{figure}

This paper focuses on peduncle detection in order to improve autonomous harvesting. 
It is highly desirable to be able to identify the peduncle prior to performing crop removal (e.g., cutting the peduncle or pulling of the crop using a custom end-effector) as this is one of the most challenging steps of the harvesting process.
Accurate estimation of the peduncle reduces the potential of damaging the crop and other branches of the plant.  Additionally, retaining the proper peduncle maximises the storage life and market value of each pepper. More importantly, accurate peduncle detection can lead to higher success rates for crop detachment, which in turn yields more harvested crops.
Peduncle detection is challenging because of varying lighting conditions and the presence of occlusions by leaves or other crops as shown in Fig.~\ref{fig:first_image}. Additionally, while peduncles are usually green, crops such as sweet peppers vary in colour from green through to red (with other variations possible), making it difficult to detect peduncles based on colour information alone.

Prior work in peduncle detection has either identified the peduncle post-harvest or yielded accuracies that are too low to be of practical use.
In~\cite{blasco2003machine, Ruiz:1996aa} the detection of the peduncle was performed using a Bayesian discriminant model of RGB colour information.
However, such an approach is not practical in-situ for crops such as sweet pepper where the colour alone does not discriminate between peduncles, leaves and crops (see Fig.~\ref{fig:first_image}).
In-situ detection of peduncle (multi-class detection) using multi-spectral imagery was performed for sweet peppers by Bac \textit{et al.}~\cite{Bac2013a}. Unfortunately their accuracy was too low to be of practical use.
In order to address previous shortcomings, we use both colour and geometry shape features captured from multiple views by an RGB-D camera. 

The aim and scope of this paper is peduncle detection from a 3D reconstructed model of a detected sweet pepper. In our previous work~\cite{sa2016deepfruits}, we described a detection system based on deep neural networks which is capable of performing the detection step and in~\cite{Lehnert:2016aa}, we introduced a sweet pepper pose estimation method that combines point clouds from multiple viewpoints into a coherent point cloud using Kinect Fusion~\cite{newcombe2011kinectfusion}. In this work, we thus assume that pre-registered 3D models of the scene containing the peduncle and sweet pepper are already obtained and the presented method is deployed afterwards.

The main contribution of this paper is the use of 3D geometric features, namely Point Feature Histograms (PFH)~\cite{rusu2009fast} in addition to colour information for robust and accurate peduncle detection. The proposed detection system operates with any sweet pepper type (red, green, and mixed), and achieves high accuracy.
To encourage reproducibility, we share the manually annotated set of 3D capsicum and peduncle models along with their annotated ground truth used to generate the results in this paper\footnote{\label{foot:video_data} Available at: \url{http://goo.gl/8BtcQX}}.  

The remainder of this paper is structured as follows. Section \ref{sec:background} introduces relevant related works. Section \ref{sec:detection_algorithm} describes sweet pepper peduncle detection and Section \ref{sec:data_collection} elaborates data collection environments and procedures. We present our experimental results in Section \ref{sec:results}. Conclusions are drawn in Section \ref{sec:conclusion}.

\section{Related Work/Background}\label{sec:background}

This section reviews existing methods for detecting peduncles and crops using 3D geometry and visual features. Such techniques are widely used for autonomous crop inspection and detection tasks. Cubero \textit{et al.} demonstrated the detection of various fruit peduncles using radius and curvature signatures \cite{Cubero201462}. The Euclidean distance and the angle rate change between each of the points on the contour and the fruit centroid are calculated. The presence of peduncles yields rapid changes in these metrics and can be detected using a specified threshold. Blasco \textit{et al.} \cite{blasco2003machine} and Ruiz \textit{et al.}\cite{Ruiz:1996aa} presented peduncle detection of oranges, peaches, and apples using a Bayesian discriminant model of RGB colour information. The size of a colour segmented area was then calculated and assigned to pre-defined classes. These methods are more likely suitable for the quality control and inspection of crop peduncles after the crop have been harvested rather than for harvesting automation, as they require an inspection chamber that provides ideal lighting conditions with a clean background, no occlusions, good viewpoints, and high-quality static imagery. 

There has also been significant progress in crop detection and plant classification for in-field crop harvesting \cite{dey2012classification}. Recently, a similar work for autonomous sweet pepper harvesting in greenhouses was presented by \cite{hemming2014robot}. The harvesting method within this project used only the centroid of each sweet pepper, with a harvesting tool that encloses the fruit and cuts the peduncle without its detection. Within this project, multi-spectral images (robust under varying lighting conditions) were tested to classify different parts of sweet pepper plants \cite{Bac2013a}. This work achieved 40\% multi-classification for parts of the plant. Although these results were deemed insufficient for use on a real robot, they demonstrated significant progress towards a plant classification system based on hyper-spectral information.

Using an accurate laser scanner would be beneficial for plant classification. Paulus \textit{et al.} \cite{Paulus2013} showed that the point feature histograms obtained from different plant structures, such as leaves and stems, are highly distinguishable. They, in turn, could easily classify plant organs with high accuracy using a supervised classification algorithm. Wahabzada \textit{et al.} \cite{Wahabzada2015} demonstrated grapevine, wheat, and barley organ classification using data acquired from a 2D laser scanner mounted on a robotic arm. A 3D point cloud with an accuracy of 45\unit{um} was produced from the scan using a $k$-means clustering algorithm. Our work differs from this approach in that we use a consumer-level affordable RGB-D camera with significant measurement noise and apply it in a field site.

Strawberry peduncle detection was reported by \cite{Hayashi:2010aa}. The region of interest (ROI) was pre-defined using prior knowledge and the boundary point between a fruit and peduncle was detected using colour information. The inclination of the peduncle - the angle between the vertical axis and the boundary point - was computed. It is possible to easily distinguish the boundary point for red strawberry but is a challenge to exploit this approach for green sweet pepper. 

\begin{figure}
\begin{center}
\subfigure[]{\includegraphics[height=0.4\columnwidth]{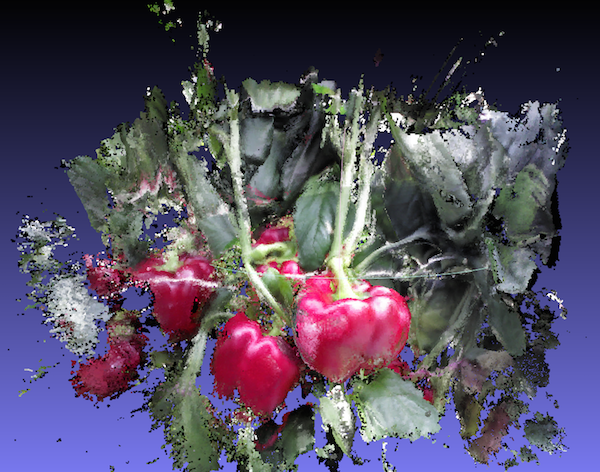}}
\subfigure[]{\includegraphics[height=0.4\columnwidth]{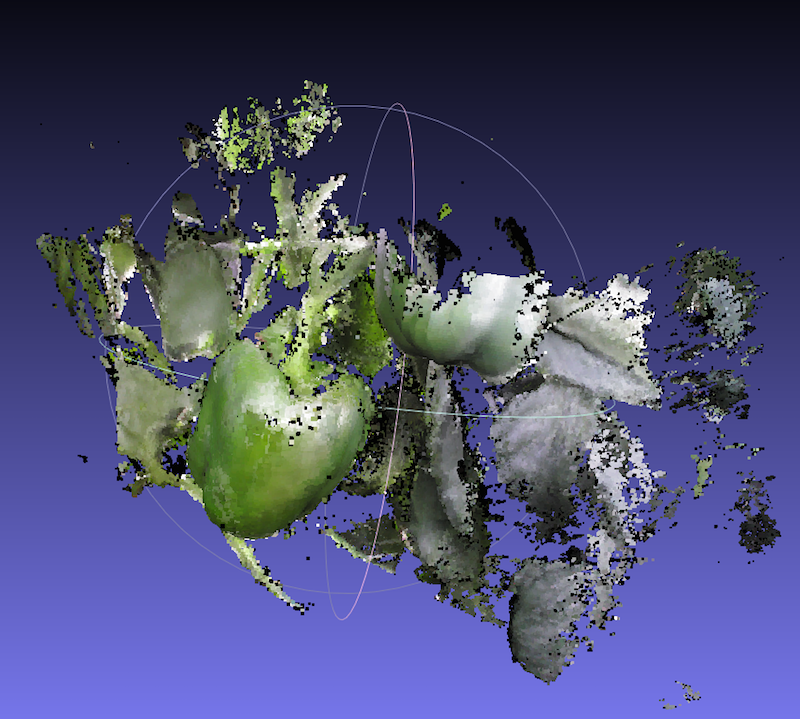}}
\end{center}
	\caption{Reconstructed red (a) and green (b) sweet peppers using Kinect Fusion. The texture and colour information is projected onto the 3D model as an average weighting from multiple viewpoints of an RGB-D camera.}
	\label{fig:3D_models}
\end{figure}

\section{Methodology}\label{sec:detection_algorithm}
Detecting the peduncles of sweet peppers is a challenging task. Unlike other crops, such as apples and mangoes, which have straight and vertical peduncles, the peduncles of sweet peppers are highly curved and sometimes even flattened against the top of the fruit as shown in Fig.~\ref{fig:first_image} and Fig.~\ref{fig:3D_models}. 

In our work, we combine colour information that can distinguish red sweet peppers from their green peduncles with geometry features~\cite{rusu2009detecting} that discriminate between peppers and peduncles that are both green. In this Section, we present a feature vector representation and the use of a supervised classifier for peduncle detection. 
\vspace{-2pt}

\subsection{Feature representation}
It is important to determine the discriminative features of an object we wish to detect. To achieve this, we utilise two features: colour and geometry shape.
\subsubsection{Colour feature (HSV)}
As shown in Fig.~\ref{fig:3D_models} (a), a highly distinguishable feature between a red sweet pepper and its peduncle is colour. Selecting a colour space for object detection is non-trivial \cite{tkalcic2003colour} and we use the Hue, Saturation, and Value (HSV) colour space that expresses visual information in the dimensions of colour (H), lightness (S) and intensity (V). Since the HSV space has a component accounting for pixel brightness, its colour components are not significantly affected by varying shades of the same colour. This is useful for visual crop detection, given that 3D objects reflect light differently depending on the angle of incidence onto a surface as shown in Fig.~\ref{fig:3D_models} (a). 
Furthermore, using the RGB colour space may not be appropriate for peduncle detection due to the high correlation between its R, G, and B dimensions. This can result in problems under certain conditions where light reflects off differently from solid-colour objects and shadows obscured by the plants. 

\subsubsection{Geometry feature (Point feature histograms \cite{rusu2009fast})}
Although colour cues provide good distinguishable features for detecting red sweet pepper peduncles, they are only of limited use for green peppers as shown in Fig.~\ref{fig:3D_models} (b) since the fruit and its peduncle have similar colour responses. Furthermore, using only colour, it is difficult to distinguish between green peduncles and other plant parts, such as leaves or stems, which are also green.

\begin{figure}
\begin{center}
\includegraphics[width=0.9\columnwidth]{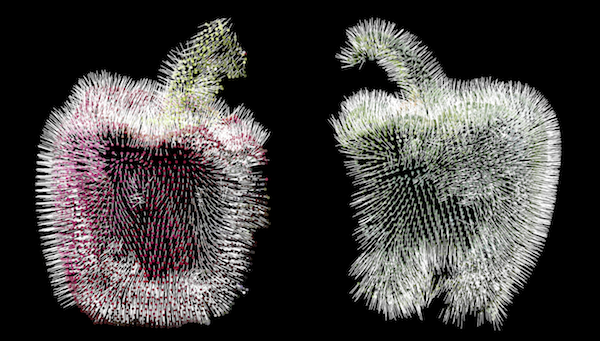}
\end{center}
	\caption{Surface normal visualisation for a red (left) and a green (right) sweet pepper. This step is important to inspect and select the right search radius $r_{\mbox{\tiny{N}}}$ for calculating point normals. This $r_{\mbox{\tiny{N}}}$ should be properly chosen based on the level of detail required by the application. In this paper, we choose $r_{\mbox{\tiny{N}}}$ as 0.01\unit{m} in order to capture the curvature of a sweet pepper body and a peduncle.}
	\label{fig:surf_normal}
\vspace{-5pt}
\end{figure}

Using geometry features, such as surface normal and curvature estimates, can be useful for identifying different classes with distinguishable 3D geometries. Fig~\ref{fig:surf_normal} shows surface normals of red (left) and green (right) sweet peppers. Here, it can be qualitatively observed that the peduncle and the body have different curvature and surface normal directions. In other words, the curvature radius for a peduncle can be generally easily distinguished from that of a sweet pepper body.
\begin{figure}[t]
\begin{center}
\subfigure[]{\includegraphics[height=0.25\columnwidth]{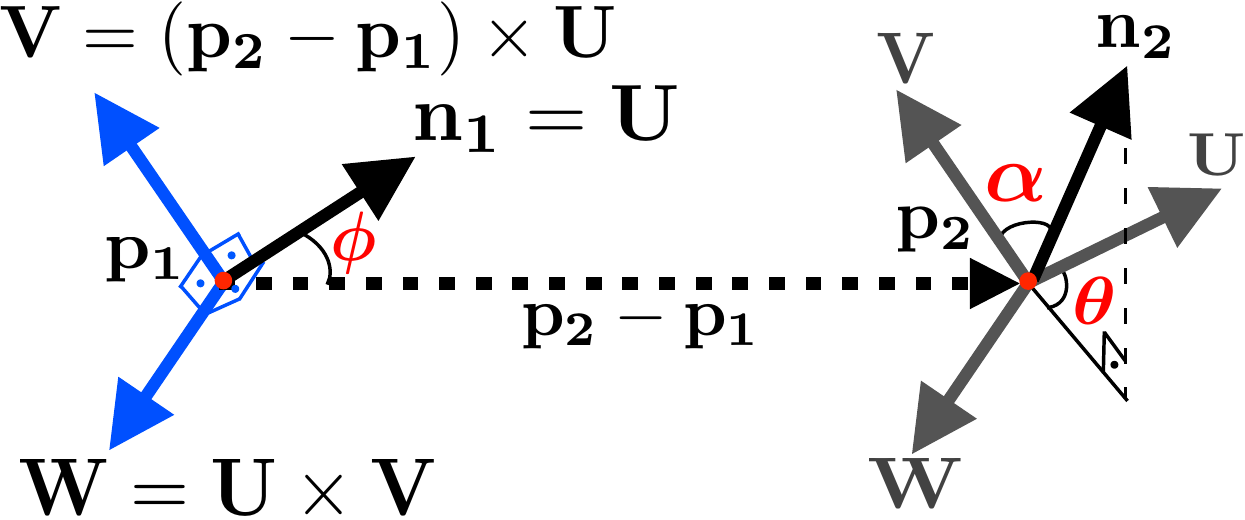}}\;\;
\subfigure[]{\includegraphics[height=0.25\columnwidth]{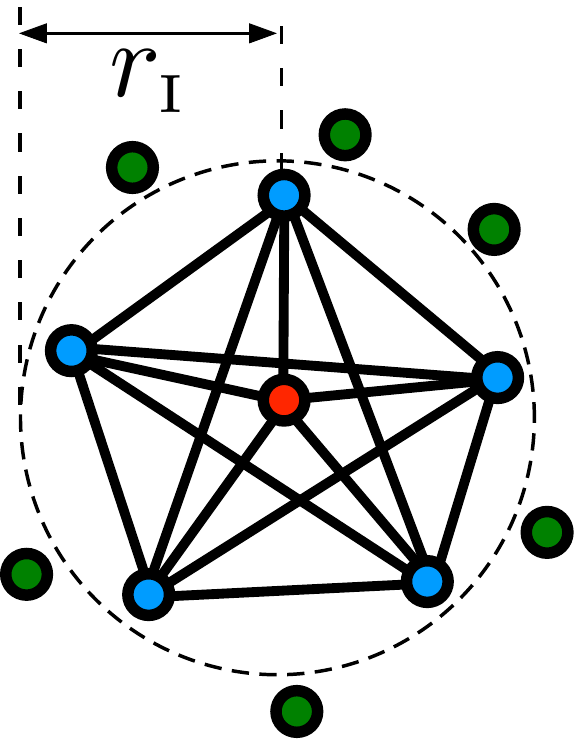}}
\end{center}
\vspace{-10pt}
	\caption{(a) Illustration of three estimated angular features for a pair of points $\bold{p_1}$ and $\bold{p_2}$ and their associated normals $\bold{n_1}$ and $\bold{n_2}$ \cite{rusu2009fast}. (b) The influence region $r_{\mbox{\tiny{I}}}$ of a Point Feature Histogram. The red and green indicates the query point and its $k$-neighbours respectively.}
	\label{fig:pfh}
\vspace{-15pt}
\end{figure}

Point Feature Histograms (PFHs) \cite{rusu2009fast} are an example of a descriptor that can capture geometry characteristics from a 3D point cloud. They represent primary surface properties given a sample point $\bold{p}$ by combining geometrical information such as Euclidean distance and normals between the query point $p$ and its $k$ neighbours as shown in Fig.~\ref{fig:pfh}(b). Given two points $\bold{p_1}$ and $\bold{p_2}$ from Fig.~\ref{fig:pfh}(a) and their normals $\bold{n_1}$ and $\bold{n_2}$, the quadruplet consisting of $<\alpha,\phi,\theta,d>$ can be formulated as:
\begin{align}
\notag
\alpha=&\bold{V}\cdot\bold{n_2}\\
\phi=&\bold{U}\cdot\frac{(\bold{p_2}-\bold{p_1})}{d}\\
\notag
\theta=&\unit{arctan}(\bold{W}\cdot\bold{n_2}, \bold{U}\cdot\bold{n_2})\\
d=&\left\Vert\bold{p}_2-\bold{p}_1\right\Vert_2
\notag
\label{eq:pfh}
\end{align}
where $\bold{U}$, $\bold{V}$, and $\bold{W}$ determine an oriented orthonormal basis of the curve (Darboux frame) shown in Fig.~\ref{fig:pfh}(a).
The quadruplets are calculated for each pair of points within the influence region $r_{\mbox{\tiny{I}}}$ (the dotted line from Fig.~\ref{fig:pfh}(b)) and then binned into a histogram of size 33 \cite{aldoma2012point} that divides the feature space.
It is important to mention that the performance of feature extraction is affected by parameters such as the normal search radius and the influence region radius as illustrated in Fig.~\ref{fig:pfh}(b). These parameters should be appropriately selected depending on the application. If these values are set too large, then the surface normals of a peduncle and body will be identical since the set of points within the radius cover points from neighbouring surfaces. However, if these values are too small, then the normal will be dominated by random noise and will not properly represent surface curvature and geometry shape. In this paper, we set $r_{\mbox{\tiny{I}}}$, and $r_{\mbox{\tiny{N}}}$= 0.01 based on empirical evaluations such as those shown in Fig.~\ref{fig:surf_normal}.

\subsection{Support Vector Machines (SVMs) classifier tuning}
At this stage, we can build a feature vector for a single 3D point with 36 float elements extracted from HSV (3) and PFH (33). A concatenated feature vector $(\unit{n}\times36)$ where $\unit{n}$ is the number of points, serves as an input for a support vector machine (SVM) classifier. SVMs are one of the most successful and popular binary supervised learning algorithms (i.e. two-class classification) due to their simplicity, flexibility, effectiveness in dealing with high-dimensional spaces and ease of training. We provide insight into our experimental datasets in the following section, as well as choosing a kernel function and its parameters in section \ref{sec:tuning}. A more in-depth  explanation of SVMs can be found in~\cite{bishop2006pattern} chapter 4.

\begin{figure}
\begin{center}
\includegraphics[width=\columnwidth]{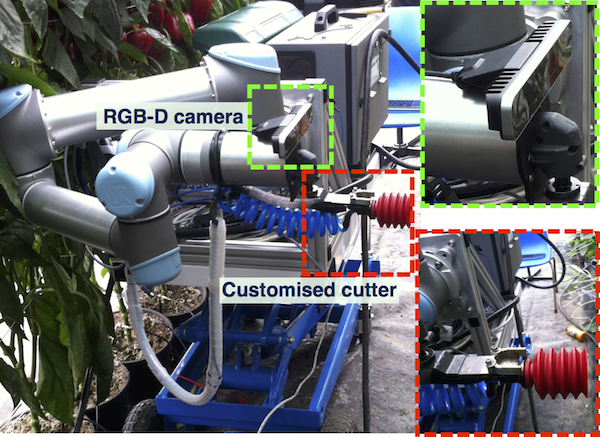}
\end{center}
\vspace{-15pt}
	\caption{QUT's Harvey, a sweet pepper harvester prototype. In this paper, we only use the robotic arm and RGB-D sensor for peduncle detection. System integration with the proposed peduncle detection and the custom end-effector to perform autonomous sweet pepper harvesting will be tackled in future work.}
	\label{fig:harvey_pic}
\vspace{-10pt}
\end{figure}
\section{System setup and data collection}\label{sec:data_collection}
Image data training and system testing was performed using a prototype robotic sweet pepper harvester, ``Harvey", we developed at QUT. As shown in Fig.~\ref{fig:harvey_pic}, this device possesses a 6 DoF robotic arm from Universal Robots (UR5) mounted on a scissor lift platform. The robot arm has a customised harvesting tool mounted on its end effector that both grips each pepper with a suction cup and cuts the peduncle with an oscillating cutting blade. An Intel i7 PC records incoming RGB-D data (640$\times$480) from a eye-in-hand depth camera\footnote{\label{foot:rgbd_camera}Intel Real Sense F200, shorter range (0.2\unit{m}-1.2\unit{m})}.

\begin{figure}
\begin{center}
\includegraphics[width=\columnwidth]{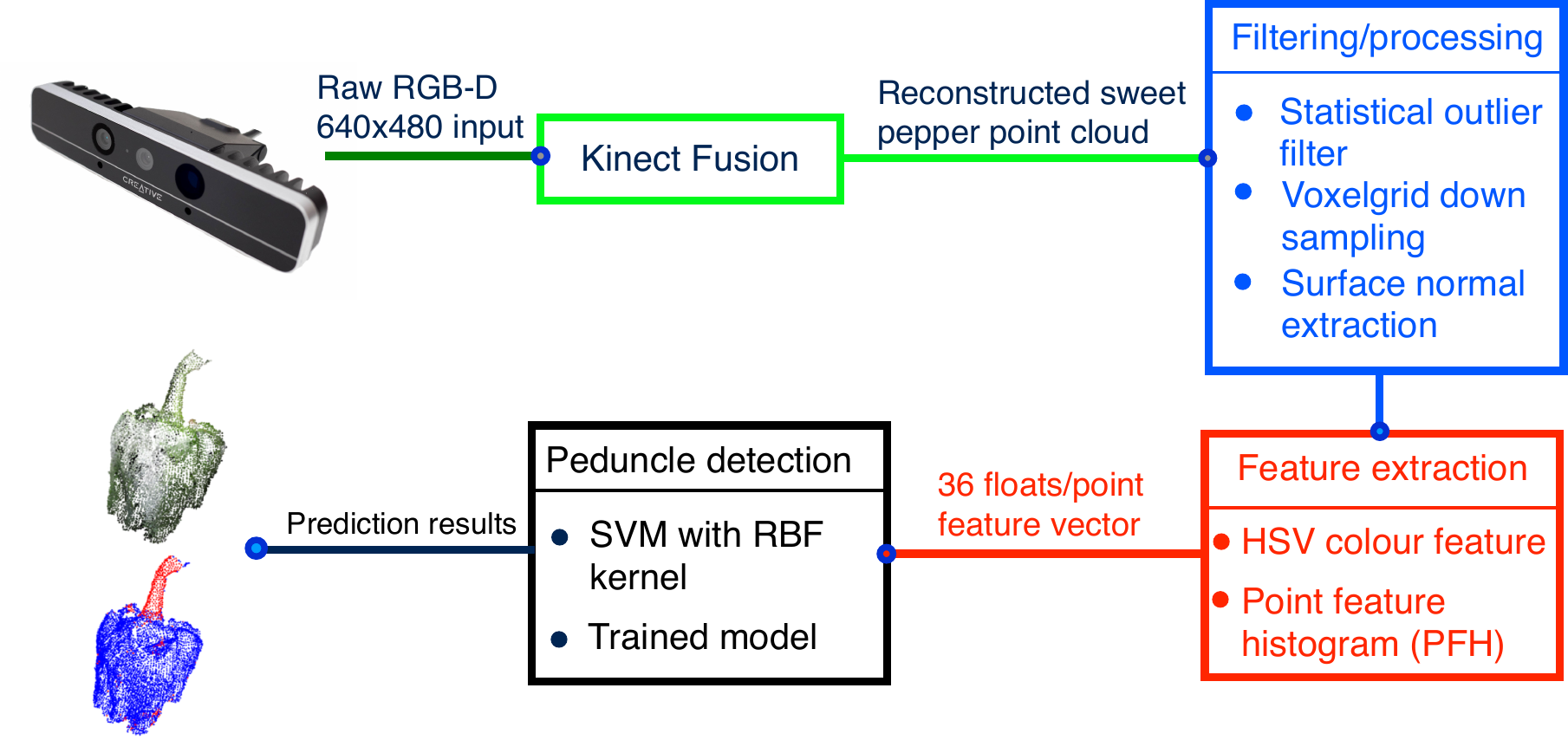}
\end{center}
\vspace{-15pt}
	\caption{Software system pipeline for the peduncle detection. Each box denotes a ROS node and its output.}
	\label{fig:pipe_line}
\vspace{-5pt}
\end{figure}

Fig.~\ref{fig:pipe_line} shows the internal peduncle detection pipeline. The reconstruction of a dense sweet pepper point cloud from multiple views using Kinect fusion is implemented based on our previous work~\cite{Lehnert:2016aa} (green). The point clouds are extremely large and noisy, consequently we use a statistical outlier remover and voxel grid down sampler (blue) supported from Point cloud library (PCL)~\cite{rusu20113d}. Surface normals are calculated and fed into the PFH feature extractor. The PFH features are then concatenated with HSV colour information to create the 36 dimensional feature vector (red) used for classification. Finally, peduncle detection is performed using a SVM with the RBF kernel and the trained model from Section \ref{sec:kernel_function} (black).

The Robot Operating System (ROS, Indigo) \cite{quigley2009ros} is used for system integration on an Ubuntu 14.04 Linux system. The methods in the green, blue, and red boxes shown in Fig.~\ref{fig:pipe_line} are implemented in C++ using PCL, while peduncle detection (black) is written in Python 2.7 \cite{scikit-learn}.

\subsection{Training/testing data collection}
\begin{figure*}
\begin{center}
\includegraphics[width=2\columnwidth]{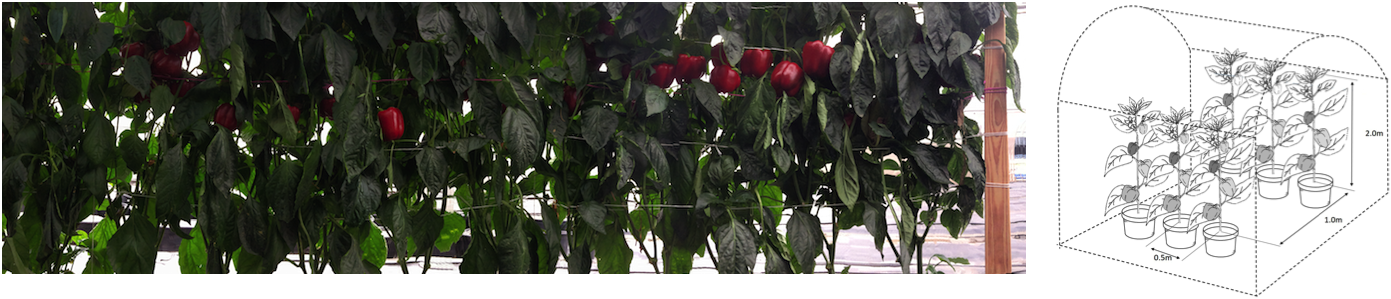}
\end{center}
\vspace{-15pt}
\caption{A panoramic view of a sweet peppers within a protected cropping facility (left) and the crop layout within it (right). It is evident that the brightness of the left and right regions is significantly different.}
\label{fig:field_photo}
\vspace{-10pt}
\end{figure*}

Field trials for data collection were conducted on a farm in North Queensland (Australia) over a 10-day period within a protected cropping system. Such a system involves growing plants under a structure (e.g., greenhouse) that aims to create an environment closer to the optimum for maximum plant growth and production as shown in Fig.~\ref{fig:field_photo}. This facility provides relatively even lighting conditions by diffusing incoming sunlight through a semi-transparent plastic/mesh roof and wall.

Data is captured while the robot arm moves through a predefined triangle scanning trajectory around the centroid of a detected red pepper in order to acquire multiple viewpoints (see time of 28\unit{s}-55\unit{s} in the demonstration video\footnoteref{foot:video_data}). The scan takes 15\unit{s}. 
During this process, not only is the red sweet pepper (in the centre) reconstructed in real-time but also the adjacent green crops. For each pepper, an RGB-D camera was moved in a scanning motion to view it from multiple directions and the Kinfu software library \cite{rusu20113d} was used to reconstruct a single point cloud.

\begin{figure}
\begin{center}
\includegraphics[width=\columnwidth]{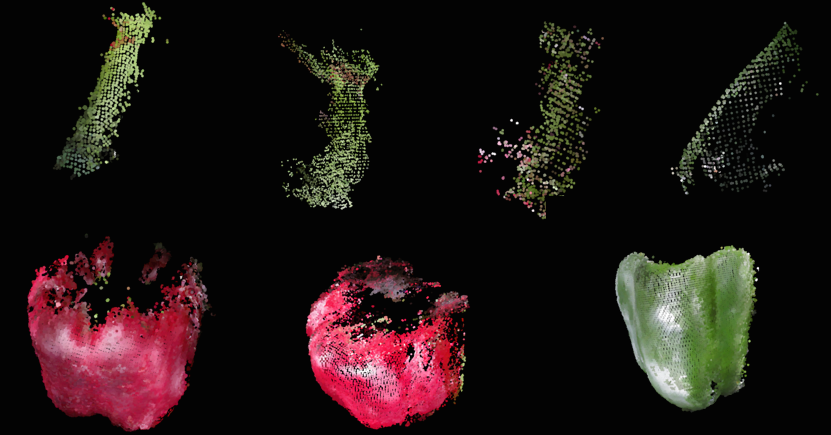}
\end{center}
\vspace{-5pt}
	\caption{Instances of manually annotated ground truth datasets for peduncles (top row) and red and green sweet peppers (bottom row). These datasets are available\protect \footnoteref{foot:video_data}.}
	\label{fig:dataset_instances}
\vspace{-10pt}
\end{figure}

We manually annotated a dataset with a total of \textbf{72} sweet peppers, example images are provided in Fig.~\ref{fig:dataset_instances}. It can be seen that these images have a substantial amount of noise present, this is due in part to the low-cost RGB-D sensor as well as shades/reflections from the object surfaces. We randomly divide these 72 samples into two even groups for model training and testing. The output from our classifier is a binary prediction for each 3D point (i.e., 0 is for sweet pepper and 1 for peduncle).

Table~\ref{tbl:dataset_anno_numbers_even} presents the first set split into a \textbf{50}-\textbf{50} ratio of training and test data whereas Table~\ref{tbl:dataset_anno_numbers_uneven} demonstrates a \textbf{79}-\textbf{21} split. These datasets are used for system evaluation in Section \ref{subsec:detection}. 

\begin{table}[h]
  \caption{Number of annotated 3D models for \textbf{50-50} ratio field trip dataset.}
  \vspace{-20pt}
  \begin{center}
    \begin{tabular}{c|c|c|c}    
    \multirow{2}{*}{}&\textbf{Train }&\textbf{Test}& \textbf{Num.} \\
    {} & (peduncle+pepper) & (peduncle+pepper) &{\textbf{3D models}} \\ \hline
    \textbf{Field Trip 1}  & 14 (50\%)    &  14 (50\%)            & 28                \\ \hline
    \textbf{Field Trip 2}  & 22 (50\%)    &  22 (50\%)            & 44              \\ \hline
    \textbf{Total}  & 36 (\textbf{50}\%) & 36 (\textbf{50}\%)              & 72         
    \end{tabular}
    \end{center}
  \label{tbl:dataset_anno_numbers_even}
  \vspace{-15pt}
\end{table}%
\begin{table}[h]
  \caption{Number of annotated 3D models for \textbf{79-21} ratio field trip dataset.}
  \vspace{-20pt}
  \begin{center}
    \begin{tabular}{c|c|c|c}    
    \multirow{2}{*}{}&\textbf{Train }&\textbf{Test}& \textbf{Num.} \\
    {} & (peduncle+pepper) & (peduncle+pepper) &{\textbf{3D models}} \\ \hline
    \textbf{Field Trip 1}  & 21 (75\%)    &  7 (25\%)            & 28                \\ \hline
    \textbf{Field Trip 2}  & 36 (82\%)    &  8 (18\%)           & 44              \\ \hline
    \textbf{Total}  & 57 (\textbf{79}\%) & 15 (\textbf{21}\%)            & 72         
    \end{tabular}
    \end{center}
  \label{tbl:dataset_anno_numbers_uneven}
 \vspace{-5pt}
\end{table}%

\begin{table}[]
\centering
\caption{Types of sweet peppers for each field trip. (The text colour indicates the sweet pepper colour, mixed=blue).}
\begin{tabular}{c|c|c|c|c}
                      & \multicolumn{2}{c|}{\textbf{Train}}    & \multicolumn{2}{c}{\textbf{Test}}   \\ \cline{2-5} 
\multirow{-2}{*}{}    & \textbf{50-50}                         & \textbf{79-21}  & \textbf{50-50}                               & \textbf{79-21} \\ \hline
\textbf{Field Trip 1} & \color{CommentRed}{\textbf{14}}, \color{CommentDG}{\textbf{0}}, \color{CommentBlue}{\textbf{0}} & \color{CommentRed}{\textbf{21}}, \color{CommentDG}{\textbf{0}}, \color{CommentBlue}{\textbf{0}} & \multicolumn{1}{c|}{\color{CommentRed}{\textbf{14}}, \color{CommentDG}{\textbf{0}}, \color{CommentBlue}{\textbf{0}}} & \color{CommentRed}{\textbf{7}}, \color{CommentDG}{\textbf{0}}, \color{CommentBlue}{\textbf{0}} \\ \hline
\textbf{Field Trip 2} & \color{CommentRed}{\textbf{18}}, \color{CommentDG}{\textbf{4}}, \color{CommentBlue}{\textbf{0}}                        & \color{CommentRed}{\textbf{29}}, \color{CommentDG}{\textbf{5}}, \color{CommentBlue}{\textbf{2}} & \multicolumn{1}{c|}{\color{CommentRed}{\textbf{15}}, \color{CommentDG}{\textbf{5}}, \color{CommentBlue}{\textbf{2}}} & \color{CommentRed}{\textbf{4}}, \color{CommentDG}{\textbf{4}}, \color{CommentBlue}{\textbf{0}}
\end{tabular}
\label{tbl:dataset_types}
\end{table}

\section{Experimental evaluation and Results}\label{sec:results}
In this section, we present qualitative and quantitative peduncle detection results. To evaluate the performance of the 3D point segmentation, we use the Area-Under-the-Curve (AUC) measure of the precision-recall curve.

\begin{figure}
\begin{center}
\subfigure[]{\includegraphics[width=0.8\columnwidth]{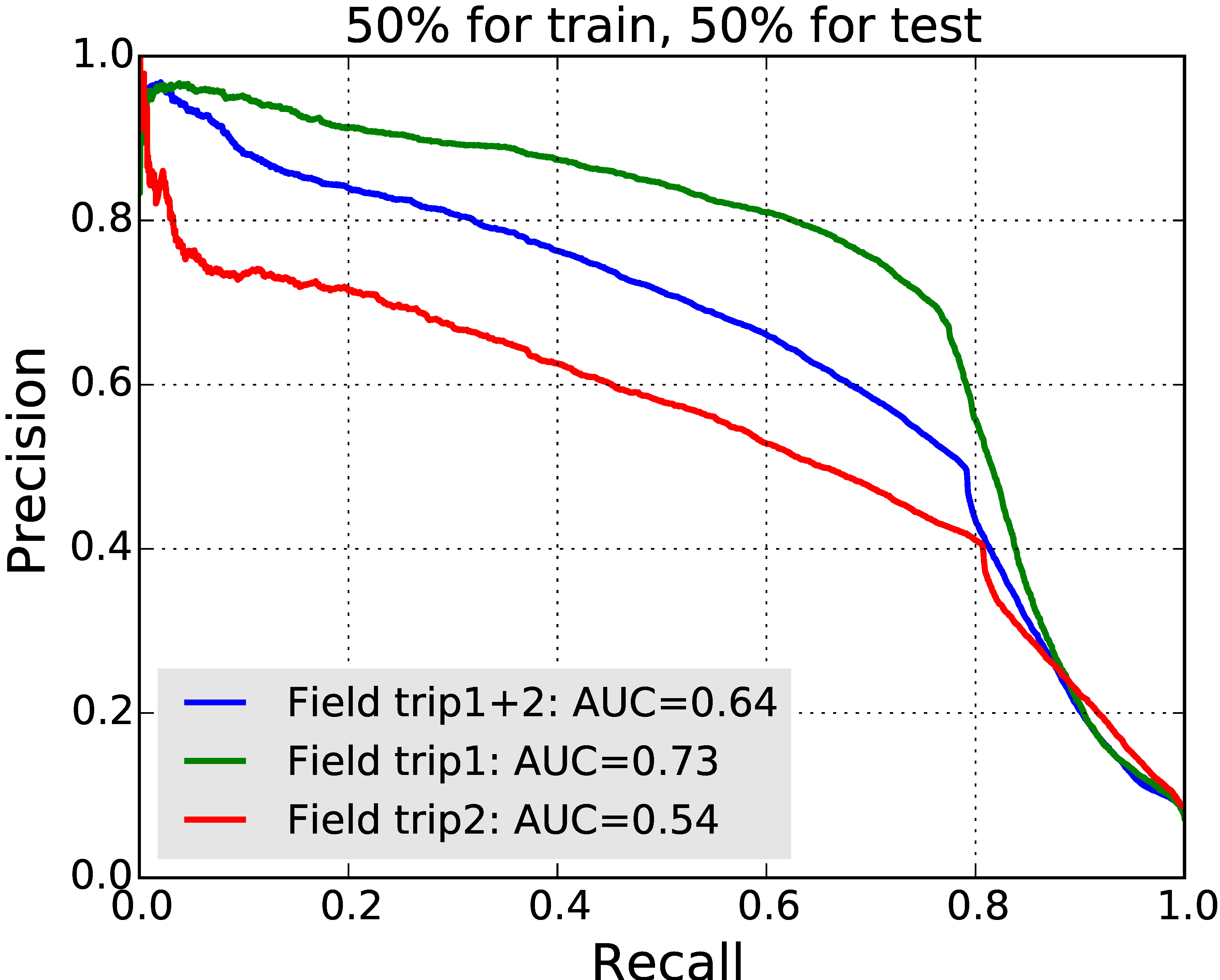}}
\subfigure[]{\includegraphics[width=0.8\columnwidth]{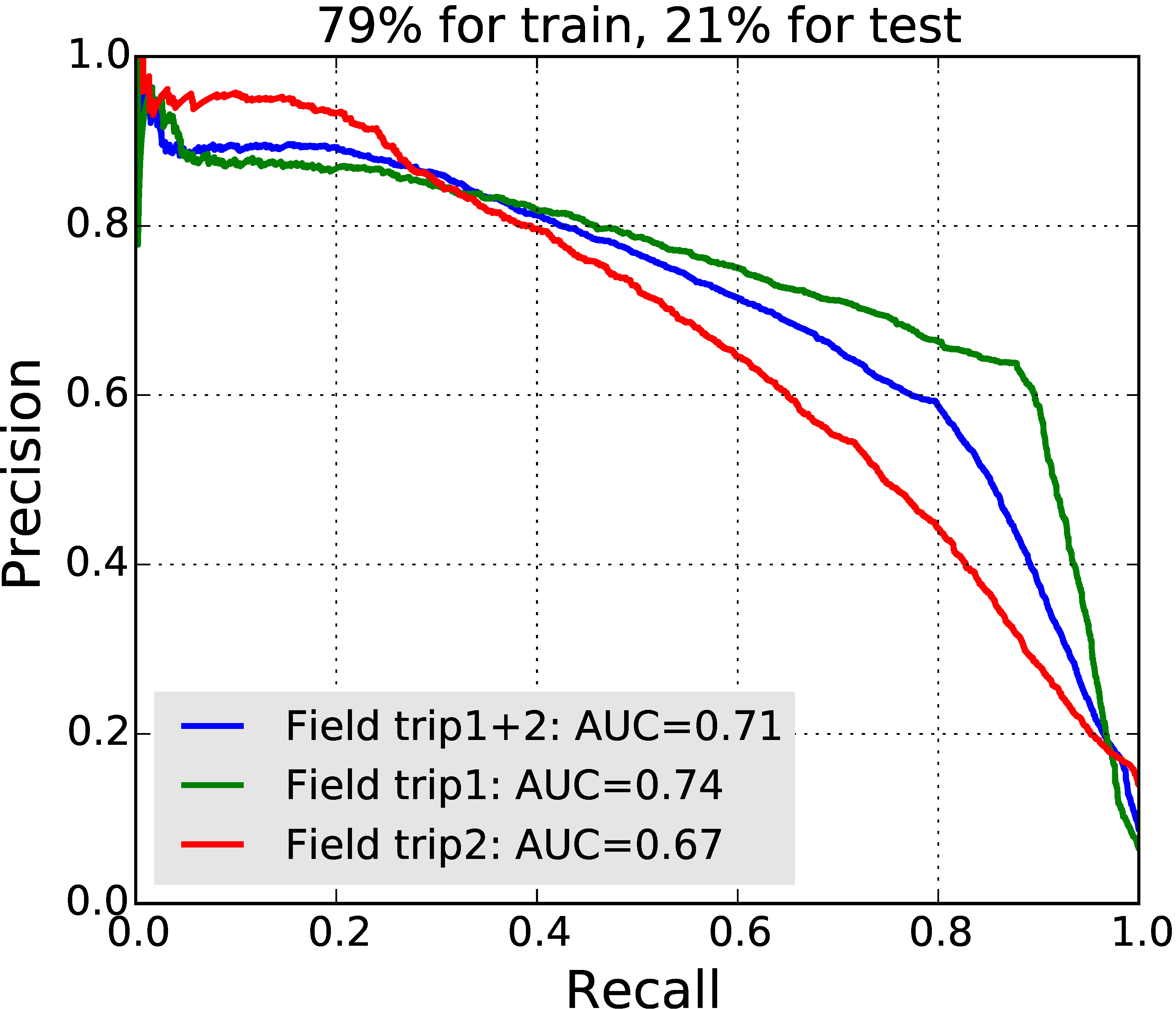}}\end{center}
\vspace{-15pt}
	\caption{Quantitative detection results of precision-recall curves and their AUC when using 50\% training and 50\% testing dataset (a) and 79\%-21\% ratio dataset (b).}
	\label{fig:detection_quatitative_results}
\vspace{-15pt}
\end{figure}

\begin{figure}[t]
\begin{center}
\includegraphics[width=0.8\columnwidth]{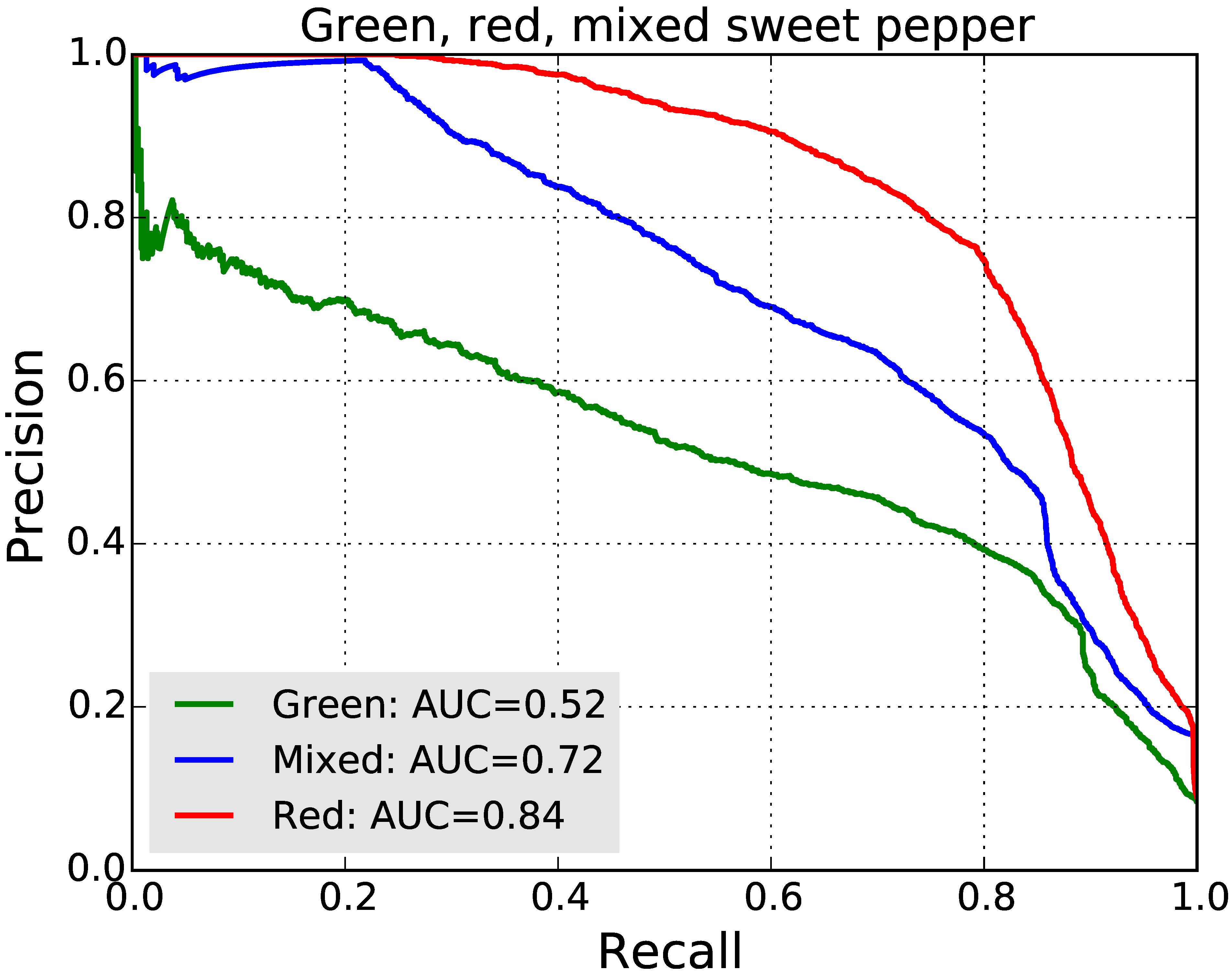}
\end{center}
\vspace{-5pt}
	\caption{Quantitative detection results of precision-recall curves and their AUC for different types of sweet peppers; green, red, and mixed colour.}
	\label{fig:green_red_mixed_AUC_results}
\vspace{-5pt}
\end{figure}

\subsection{Peduncle detection results}\label{subsec:detection}
Fig.~\ref{fig:detection_quatitative_results} shows the quantitative results of peduncle detection. We note that Fig.~\ref{fig:detection_quatitative_results}(a) uses 50\% of the available data for training and testing (Table~\ref{tbl:dataset_anno_numbers_even}) whereas  Fig.~\ref{fig:detection_quatitative_results}(b) has an uneven ratio of 79\% for training and 21\% for testing (Table~\ref{tbl:dataset_anno_numbers_uneven}). It can be seen that Field Trip 1 yields a consistent AUC result of 0.73 but that Field Trip 2 varies significantly from 0.54 to 0.67. This is mainly due to the higher number of more challenging green and mixed sweet peppers in the latter. Fig.~\ref{fig:detection_quatitative_results}(a) contains 4 green sweet peppers in the training set and 5 green and 2 mixed in the testing set.
Fig.~\ref{fig:detection_quatitative_results}(b) encapsulates 5 green and 2 mixed for training and 4 greens for testing. Table~\ref{tbl:dataset_types} summarises different types of sweet peppers (colour coded: mixed=blue) mixed in the training and testing data for various experiments in the two field trips. It is challenging to estimate the optimal ratio of training-testing datasets, especially when the number of samples is relatively low \cite{guyon1997scaling}. As a rule-of-thumb in the community, we present results with ratios of 50-50 and 80-20.

We also perform an analysis over three different types of sweet peppers red, green, and mixed colour (red and green). Using the model trained with the 50-50 ratio dataset, sweet peppers in the test dataset are clustered by their colour (red: 29, green: 5, and mixed: 2). There are two mixed sweet peppers which have different amounts of red and green colouration (red=32.5\% and 75\%, green=67.5\% and 25\% respectively). We then perform the detection shown in Fig.~\ref{fig:green_red_mixed_AUC_results}. As expected, while red sweet peppers can be easily distinguished, detection results are poorer for green ones (AUC: 0.52). This supports our previous argument on the performance drop for Field Trip 2. Mixed sweet peppers show the intermediate result of only 2 testing samples.

\subsubsection{Qualitative evaluation of detection}

\begin{figure}
\begin{center}
\includegraphics[width=\columnwidth]{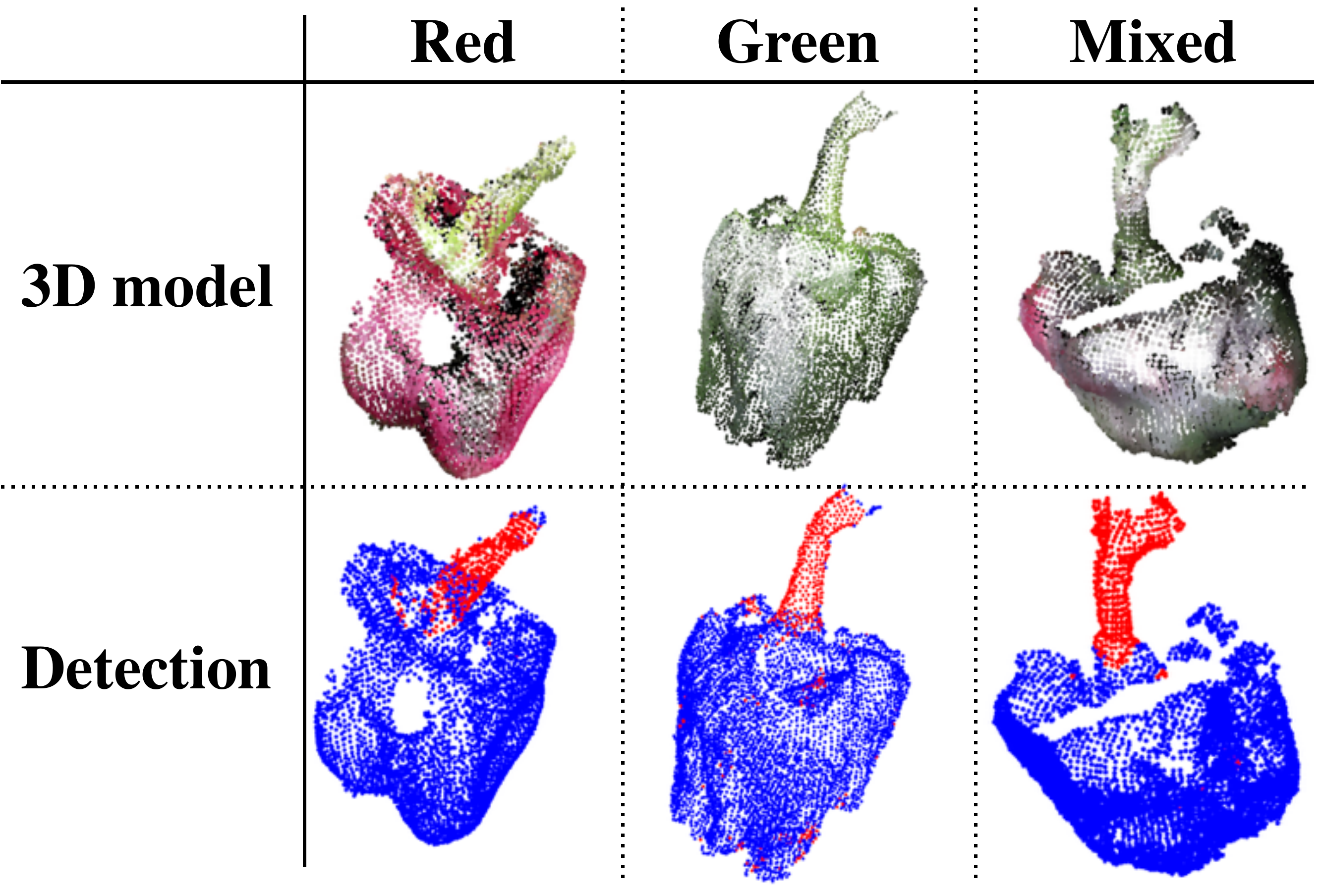}
\end{center}
\vspace{-15pt}
	\caption{Qualitative detection results. The top row indicates the input point clouds and the bottom depicts segmentation results for peduncles (red) and sweet peppers (blue). Each column represents different sweet pepper types.}
	\label{fig:detection_qualitative_results}
\vspace{-0pt}
\end{figure}

Fig.\ref{fig:detection_qualitative_results} displays the qualitative results of our sweet pepper peduncle detection method. We assess whether these outcomes can be utilised in the field. The columns indicate different types of sweet peppers and rows are colour and detection views. For red and mixed sweet peppers, the proposed algorithm reports impressive results. Although sufficient performance is achieved to detect the peduncles of green sweet peppers, there are noticeable false positives, $F_N$, that mainly stem from a small margin of sweet peppers and peduncles in colour and PFH features. This paper only presents one view point of 3D model, but the accompanying video shows entire views for further inspection.

\begin{figure}
\begin{center}
\includegraphics[width=0.8\columnwidth]{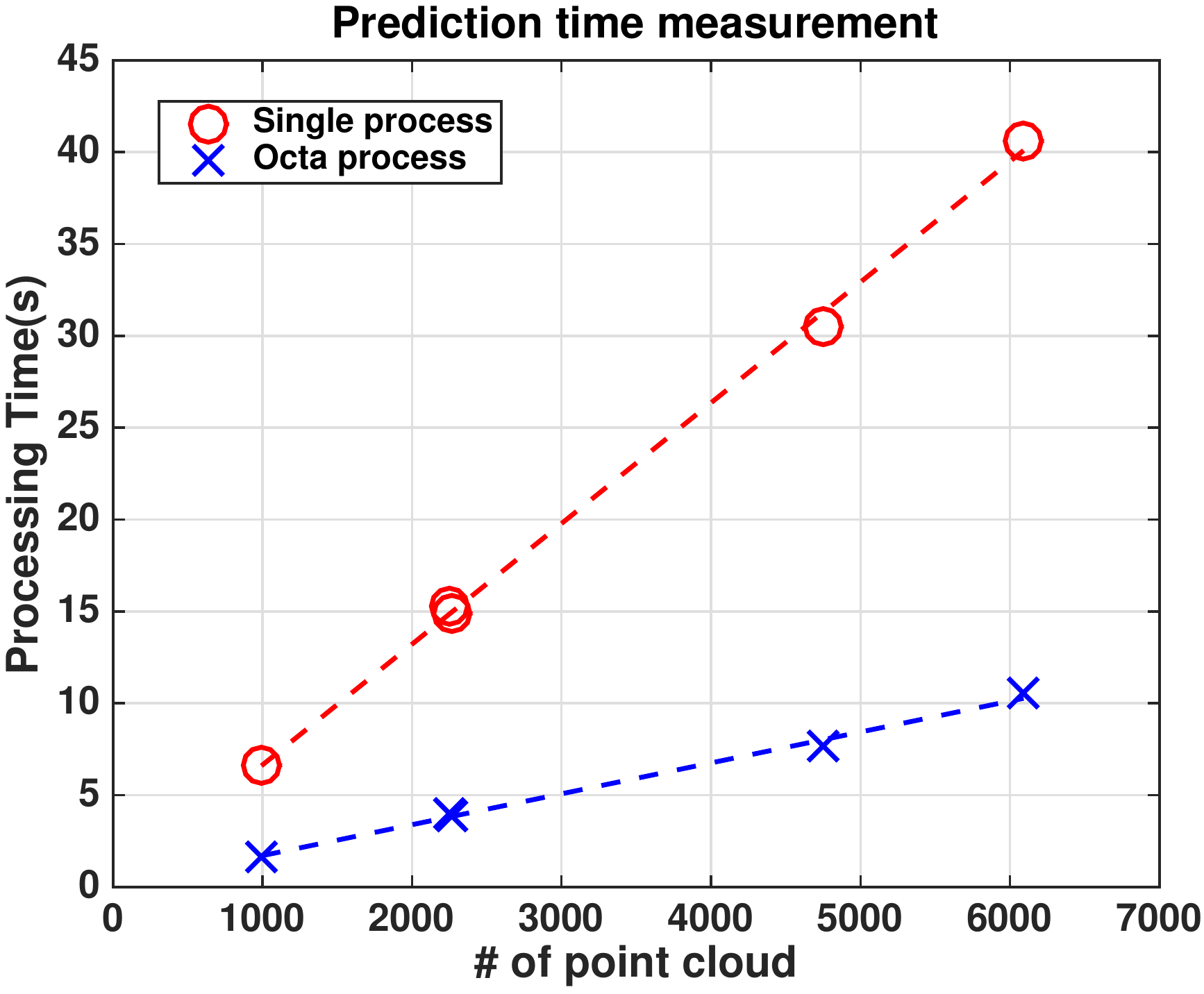}
\end{center}
\vspace{-10pt}
	\caption{Processing speed comparison for a single (red) and octa (blue) processes. A quadrouple speed improvement is achieved with parallel processing on a quad-core computer.}
	\label{fig:parallel_proc}
\vspace{-0pt}
\end{figure}
\begin{figure}
\begin{center}
\includegraphics[width=0.8\columnwidth]{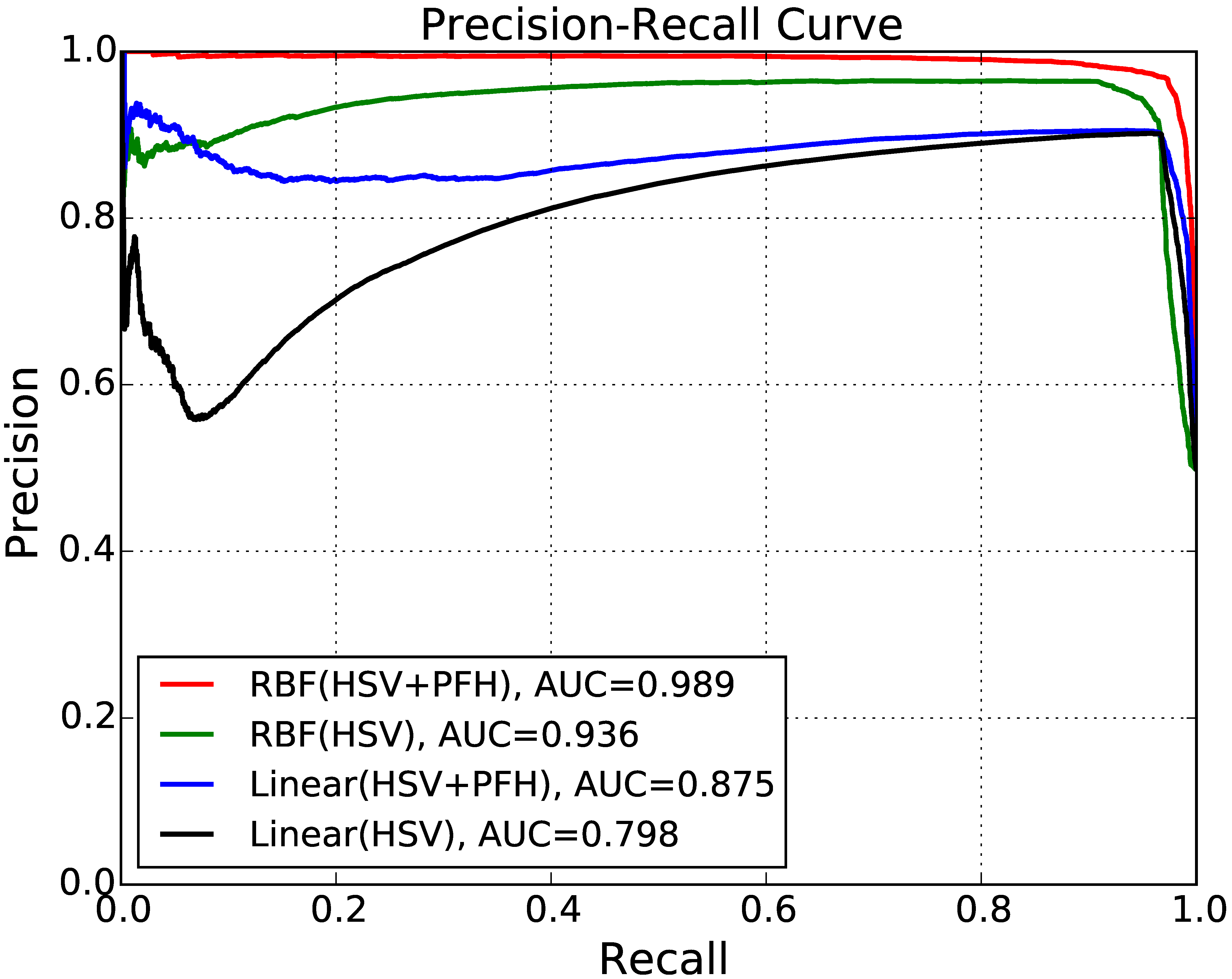}
\end{center}
\vspace{-10pt}
	\caption{The impact of PFH features using different kernels for red sweet peppers. Colour and PFH features complement each other and improve detection rate.}
	\label{fig:kernel_tuning}
\vspace{-10pt}
\end{figure}
\subsection{Processing time and parameter tuning}\label{sec:tuning}
Our testing methodology highlighted issues in processing time. For example, it takes about 40 seconds to predict the 6,094 input points, which is unlikely to be practical for autonomous sweet pepper picking. Since predicting input features is fully independent, we thus apply parallel processing, which can submit jobs across multiple processes\cite{cornelis2008fast} as shown in Fig.~\ref{fig:parallel_proc}. Feature vectors are evenly divided between the number of processes and quadrouple speed-ups can be achieved with quad cores.
\subsubsection{Model (Kernel function) selection and parameter tuning}\label{sec:kernel_function}
The aim of this study is to identify the contributions of features (i.e., colours and 3D point clouds). To achieve this, we set a dataset only with red sweet peppers and vary kernels and combinations of features. Note that the dataset used in this model selection study differ to those in Table~\ref{tbl:dataset_anno_numbers_uneven}.

We use the Machine Learning Python package \cite{scikit-learn} that supports a variety of kernel function options such as linear, polynomial, radial basis function (RBF), and even custom kernel functions. It is challenging to pick the optimal kernel and associated parameter sets for all possible combinations since training is very time-consuming (e.g., model training takes about 10 hours for linear SVMs with 50K sample points on a modern Intel i7 computer.). We thus evaluate 16 different cases for 3 common kernel functions (SVC, RBF, and linear) and 2 dominant parameters such as Gaussian kernel and regularisation terms\footnote{\url{http://scikit-learn.org/stable/auto_examples/svm/plot_rbf_parameters.html}} (named \texttt{gamma} and \texttt{C} respectively). Precision-recall curves are calculated for all cases and AUC is used as their comparison metric, with higher AUC numbers implying better detection performance. The RBF kernels with \texttt{gamma}=0.01 and \texttt{C}= 100 report the best AUC score of 0.971, where \texttt{gamma} is the inverse of the radius of a single training point accounting for hyperplane estimation and \texttt{C} defines the impact of mis-classification on the cost function. Again, we empirically choose these parameters by performing parameter sweeping (16 evaluations with varying parameters for each kernel (i.e., SVC, RBF, and linear)).

Given this optimised kernel function and parameters, we investigate the impact of each feature on peduncle detection in Fig.~\ref{fig:kernel_tuning}. We examine the impact of using just colour information (HSV, denoted as green and black in Fig.~\ref{fig:kernel_tuning}) and colour combined with geometry features (HSV+PFH, in red and blue). Moreover, we study the effects of different kernels with the same features. The plots demonstrate that RBF kernels are superior to their linear counterparts, and colour is the dominant feature for red sweet pepper detection. PFH contributes to improving the detection performance (see Fig.~\ref{fig:kernel_tuning} at recall level around 0.1). Further experimental results on different sweet peppers types, such as green and mixed colours, are presented in Section \ref{subsec:detection}.

\begin{figure}
\begin{center}
\subfigure[]{\includegraphics[width=\columnwidth]{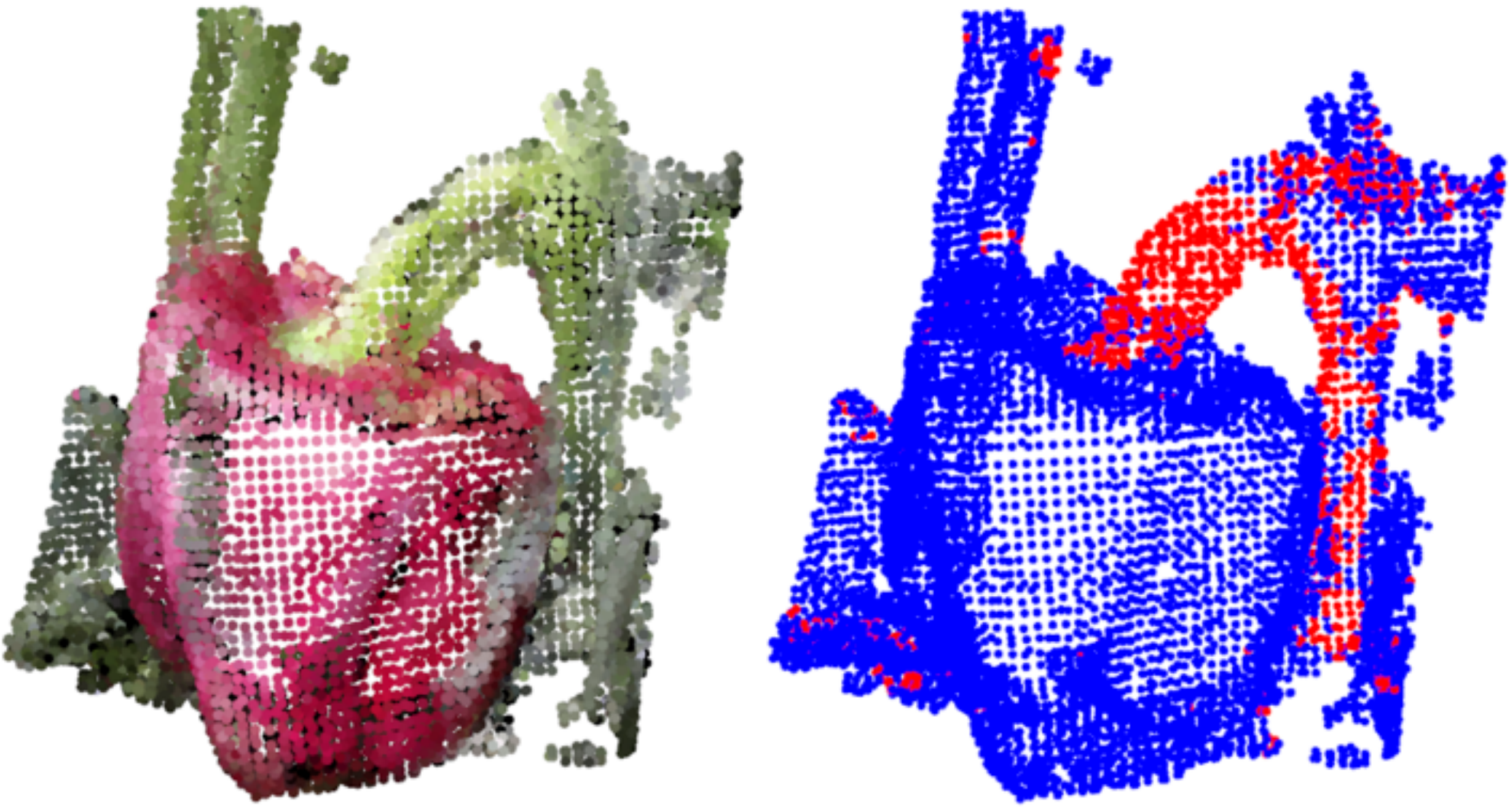}}
\subfigure[]{\includegraphics[width=\columnwidth]{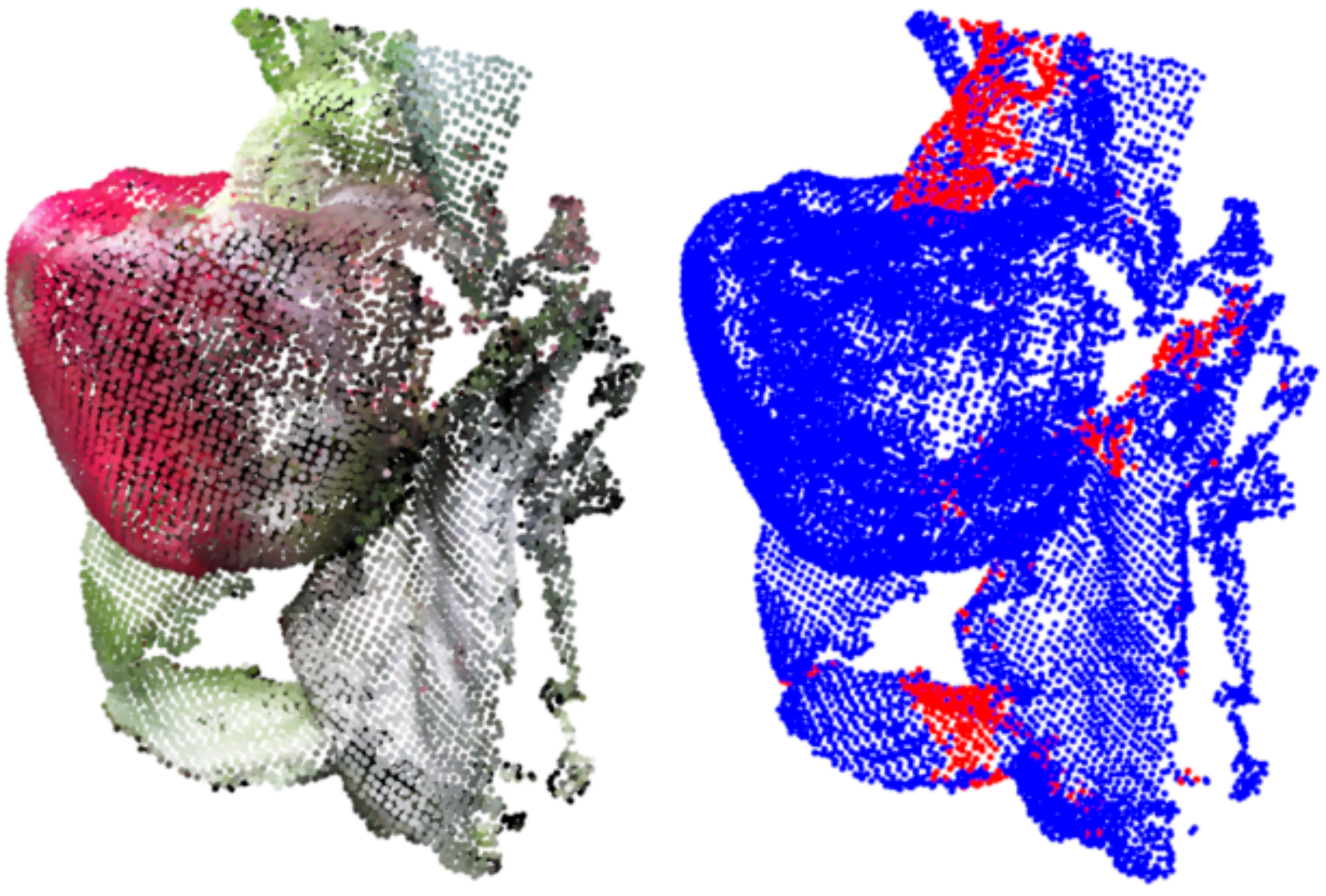}}
\subfigure[]{\includegraphics[width=\columnwidth]{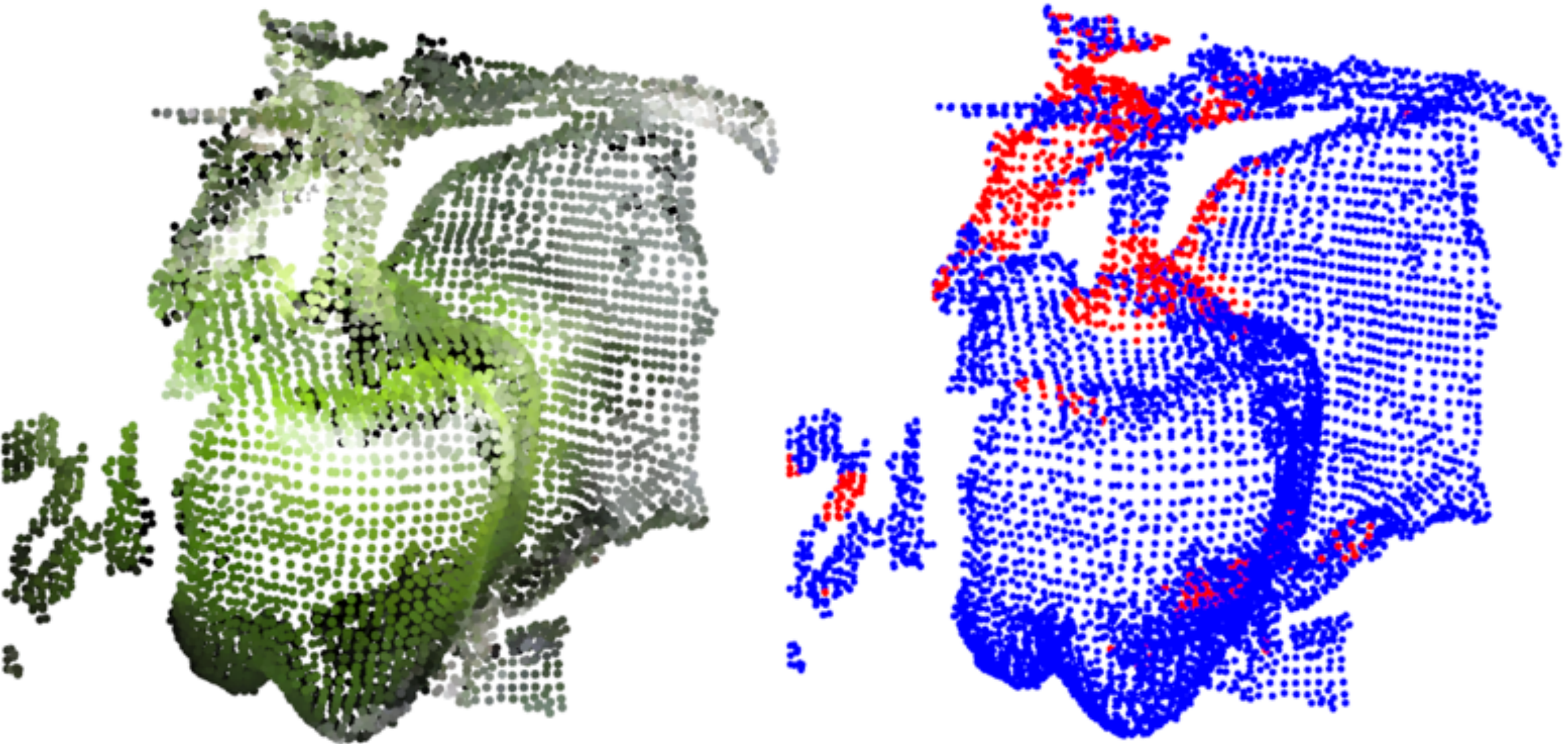}}
\end{center}
\vspace{-15pt}
	\caption{Qualitative detection results with the background presence for (a) red, (b) mixed, and (c) green sweet peppers.}
	\label{fig:detection_quatitative_results_with_background}
\vspace{-10pt}
\end{figure}

\section{Discussion}
This paper presents initial, and encouraging, results of peduncle point-level segmentation, which aims at enhancing the performance of a sweet pepper robotic harvester. The integration of the segmentation with the actual harvester and the assessment of its performance is outside the scope of the paper and part of on-going work.

A noticeable attribute of the proposed method is that the trained model yields false positives for other stems and leaves that exhibit similar colour and shape to the peduncle as shown in Fig.~\ref{fig:detection_quatitative_results_with_background}. To overcome this issue, future work could examine the potential of using context features that account for the geometrical relationship between a sweet pepper and a peduncle (i.e., a sweet pepper usually grows the direction of gravity and locates the top of the fruit). The orientation of each segmented peduncle point and its bounding volume can then be computed. With this information, the supervised learning algorithm may be able to determine that peduncles are usually above the centroid within a range of distances, reducing false positive rates. The accompanying video demonstration displays qualitative results with different viewing points\footnoteref{foot:video_data}.

\section{Conclusions and future work}\label{sec:conclusion}
We present a detection system for sweet pepper peduncles in a farm field site. Colour and shape obtained from a RGB-D sensor, are utilised as dominant features for the detection and we discuss the impact of selecting features and their tuning approaches.
 
The novel use of PFHs provides further discriminative power and demonstrates feasibility in using depth information for agricultural applications. We present promising qualitative results and quantitatively achieve an AUC of 0.71 for point-level segmentation. 

Future work will consider ways of utilising GPU parallel processing for detection speed improvement. Other classifiers such as Random Forest or Gaussian Processes will also be compared regarding their prediction speed and detection rate. With more training data, detection performance could be improved by incorporating different weather conditions in future field trips during harvesting season. Also, to avoid intensive labour work associated with manual annotation, we plan to conduct further data collection field trips in near future. We are interested in applying this system to other sweet pepper varieties or different peduncle types.

\section*{Acknowledgement}
The authors would like to thank Raymond Russell and Jason Kulk for their key contributions to the design and fabrication of the harvesting platform. We also would like to acknowledge Elio Jovicich and Heidi Wiggenhauser for their advice and support during the field trips. We thank Marija Popovi\'{c} for comments and proofreading that improved the quality of manuscript.

\bibliographystyle{IEEEtran}
\bibliography{./bibs/RAL2017}

\end{document}

%% file: extern/pic-common.tex
\def\fps@figure{htp}
\def\fps@table{htp}

\newcommand{\bi}{\begin{itemize}}
\newcommand{\ei}{\end{itemize}}

\newcommand{\bfig}{\begin{figure}}
\newcommand{\efig}{\end{figure}}

\newcommand{\benum}{\begin{enumerate}}
\newcommand{\eenum}{\end{enumerate}}

\newcommand{\be}{\begin{equation}}
\newcommand{\ee}{\end{equation}}

\newcommand{\ba}{\begin{eqnarray}}
\newcommand{\ea}{\end{eqnarray}}

\newcommand{\unit}[1]{\mbox{$\rm \,#1$}}

%% file: RAL2017_peduncle_detection.bbl
\begin{thebibliography}{10}
\providecommand{\url}[1]{#1}
\csname url@samestyle\endcsname
\providecommand{\newblock}{\relax}
\providecommand{\bibinfo}[2]{#2}
\providecommand{\BIBentrySTDinterwordspacing}{\spaceskip=0pt\relax}
\providecommand{\BIBentryALTinterwordstretchfactor}{4}
\providecommand{\BIBentryALTinterwordspacing}{\spaceskip=\fontdimen2\font plus
\BIBentryALTinterwordstretchfactor\fontdimen3\font minus
  \fontdimen4\font\relax}
\providecommand{\BIBforeignlanguage}[2]{{%
\expandafter\ifx\csname l@#1\endcsname\relax
\typeout{** WARNING: IEEEtran.bst: No hyphenation pattern has been}%
\typeout{** loaded for the language `#1'. Using the pattern for}%
\typeout{** the default language instead.}%
\else
\language=\csname l@#1\endcsname
\fi
#2}}
\providecommand{\BIBdecl}{\relax}
\BIBdecl

\bibitem{kondo2011agricultural}
N.~Kondo, M.~Monta, and N.~Noguchi, \emph{Agricultural Robots: Mechanisms and
  Practice}.\hskip 1em plus 0.5em minus 0.4em\relax Apollo Books, 2011.

\bibitem{english2014vision}
A.~English, P.~Ross, D.~Ball, and P.~Corke, ``Vision based guidance for robot
  navigation in agriculture,'' in \emph{Robotics and Automation (ICRA), 2014
  IEEE International Conference on}.\hskip 1em plus 0.5em minus 0.4em\relax
  IEEE, 2014, pp. 1693--1698.

\bibitem{Ball:2016aa}
\BIBentryALTinterwordspacing
D.~Ball, B.~Upcroft, G.~Wyeth, P.~Corke, A.~English, P.~Ross, T.~Patten,
  R.~Fitch, S.~Sukkarieh, and A.~Bate, ``Vision-based obstacle detection and
  navigation for an agricultural robot,'' \emph{Journal of Field Robotics}, pp.
  n/a--n/a, 2016. [Online]. Available:
  \url{http://dx.doi.org/10.1002/rob.21644}
\BIBentrySTDinterwordspacing

\bibitem{sa2015visual}
I.~Sa, C.~McCool, C.~Lehnert, and T.~Perez, ``On visual detection of
  highly-occluded objects for harvesting automation in horticulture,'' in
  \emph{IEEE International Conference on Robotics and Automation, Workshop on
  Robotics in Agriculture}.\hskip 1em plus 0.5em minus 0.4em\relax ICRA, 2015.

\bibitem{mccool2016visual}
C.~McCool, I.~Sa, F.~Dayoub, C.~Lehnert, T.~Perez, and B.~Upcroft, ``Visual
  detection of occluded crop: For automated harvesting,'' in \emph{2016 IEEE
  International Conference on Robotics and Automation (ICRA)}.\hskip 1em plus
  0.5em minus 0.4em\relax IEEE, 2016, pp. 2506--2512.

\bibitem{Lehnert:2016aa}
C.~Lehnert, I.~Sa, C.~McCool, B.~Upcroft, and T.~Perez, ``{Sweet Pepper Pose
  Detection and Grasping for Automated Crop Harvesting},'' in \emph{IEEE
  International Conference on Robotics and Automation}, 2016.

\bibitem{bac2014harvesting}
C.~W. Bac, E.~J. Henten, J.~Hemming, and Y.~Edan, ``{Harvesting Robots for
  High-value Crops: State-of-the-art Review and Challenges Ahead},''
  \emph{Journal of Field Robotics}, vol.~31, no.~6, pp. 888--911, 2014.

\bibitem{Kitamura:2005aa}
S.~Kitamura, K.~Oka, and F.~Takeda, ``Development of picking robot in
  greenhouse horticulture,'' in \emph{Proceedings of the SICE Annual
  Conference, Okayama, Aug}.\hskip 1em plus 0.5em minus 0.4em\relax Citeseer,
  2005, pp. 8--10.

\bibitem{blasco2003machine}
J.~Blasco, N.~Aleixos, and E.~Molt{\'o}, ``Machine vision system for automatic
  quality grading of fruit,'' \emph{Biosystems Engineering}, vol.~85, no.~4,
  pp. 415--423, 2003.

\bibitem{Ruiz:1996aa}
L.~A. Ruiz, E.~Molt{\'o}, F.~Juste, F.~Pl{\'a}, and R.~Valiente, ``Location and
  characterization of the stem--calyx area on oranges by computer vision,''
  \emph{Journal of Agricultural Engineering Research}, vol.~64, no.~3, pp.
  165--172, 1996.

\bibitem{Bac2013a}
\BIBentryALTinterwordspacing
C.~Bac, J.~Hemming, and E.~van Henten, ``{Robust pixel-based classification of
  obstacles for robotic harvesting of sweet-pepper},'' \emph{Computers and
  Electronics in Agriculture}, vol.~96, pp. 148--162, 2013. [Online].
  Available:
  \url{http://www.sciencedirect.com/science/article/pii/S0168169913001099}
\BIBentrySTDinterwordspacing

\bibitem{sa2016deepfruits}
I.~Sa, Z.~Ge, F.~Dayoub, B.~Upcroft, T.~Perez, and C.~McCool, ``Deepfruits: A
  fruit detection system using deep neural networks,'' \emph{Sensors}, vol.~16,
  no.~8, p. 1222, 2016.

\bibitem{newcombe2011kinectfusion}
R.~A. Newcombe, S.~Izadi, O.~Hilliges, D.~Molyneaux, D.~Kim, A.~J. Davison,
  P.~Kohi, J.~Shotton, S.~Hodges, and A.~Fitzgibbon, ``Kinectfusion: Real-time
  dense surface mapping and tracking,'' in \emph{Mixed and augmented reality
  (ISMAR), 2011 10th IEEE international symposium on}.\hskip 1em plus 0.5em
  minus 0.4em\relax IEEE, 2011, pp. 127--136.

\bibitem{rusu2009fast}
R.~B. Rusu, N.~Blodow, and M.~Beetz, ``{Fast point feature histograms (FPFH)
  for 3D registration},'' in \emph{Robotics and Automation, 2009. ICRA'09. IEEE
  International Conference on}.\hskip 1em plus 0.5em minus 0.4em\relax IEEE,
  2009, pp. 3212--3217.

\bibitem{Cubero201462}
\BIBentryALTinterwordspacing
S.~Cubero, M.~P. Diago, J.~Blasco, J.~Tard{\'a}guila, B.~Mill{\'a}n, and
  N.~Aleixos, ``A new method for pedicel/peduncle detection and size assessment
  of grapevine berries and other fruits by image analysis,'' \emph{Biosystems
  Engineering}, vol. 117, pp. 62 -- 72, 2014, image Analysis in Agriculture.
  [Online]. Available:
  \url{http://www.sciencedirect.com/science/article/pii/S1537511013000950}
\BIBentrySTDinterwordspacing

\bibitem{dey2012classification}
D.~Dey, L.~Mummert, and R.~Sukthankar, ``Classification of plant structures
  from uncalibrated image sequences,'' in \emph{Applications of Computer Vision
  (WACV), 2012 IEEE Workshop on}.\hskip 1em plus 0.5em minus 0.4em\relax IEEE,
  2012, pp. 329--336.

\bibitem{hemming2014robot}
J.~Hemming, C.~Bac, B.~van Tuijl, R.~Barth, J.~Bontsema, E.~Pekkeriet, and
  E.~Van~Henten, ``A robot for harvesting sweet-pepper in greenhouses,'' in
  \emph{Proceedings of the International Conference of Agricultural
  Engineering, 6-10 July 2014, Z{\"u}rich, Switzerland}, 2014.

\bibitem{Paulus2013}
\BIBentryALTinterwordspacing
S.~Paulus, J.~Dupuis, A.-K. Mahlein, and H.~Kuhlmann, ``{Surface feature based
  classification of plant organs from 3D laserscanned point clouds for plant
  phenotyping.}'' \emph{BMC bioinformatics}, vol.~14, p. 238, January 2013.
  [Online]. Available:
  \url{http://www.pubmedcentral.nih.gov/articlerender.fcgi?artid=3750309{\&}tool=pmcentrez{\&}rendertype=abstract}
\BIBentrySTDinterwordspacing

\bibitem{Wahabzada2015}
\BIBentryALTinterwordspacing
M.~Wahabzada, S.~Paulus, K.~Kersting, and A.-K. Mahlein, ``{Automated
  interpretation of 3D laserscanned point clouds for plant organ
  segmentation},'' \emph{BMC Bioinformatics}, vol.~16, no.~1, p. 248, 2015.
  [Online]. Available: \url{http://www.biomedcentral.com/1471-2105/16/248}
\BIBentrySTDinterwordspacing

\bibitem{Hayashi:2010aa}
S.~Hayashi, K.~Shigematsu, S.~Yamamoto, K.~Kobayashi, Y.~Kohno, J.~Kamata, and
  M.~Kurita, ``Evaluation of a strawberry-harvesting robot in a field test,''
  \emph{Biosystems Engineering}, vol. 105, no.~2, pp. 160--171, 2010.

\bibitem{rusu2009detecting}
R.~B. Rusu, A.~Holzbach, M.~Beetz, and G.~Bradski, ``Detecting and segmenting
  objects for mobile manipulation,'' in \emph{Computer Vision Workshops (ICCV
  Workshops), 2009 IEEE 12th International Conference on}.\hskip 1em plus 0.5em
  minus 0.4em\relax IEEE, 2009, pp. 47--54.

\bibitem{tkalcic2003colour}
M.~Tkalcic, J.~F. Tasic \emph{et~al.}, ``Colour spaces: perceptual, historical
  and applicational background,'' in \emph{Eurocon}, 2003, pp. 304--308.

\bibitem{aldoma2012point}
A.~Aldoma, Z.-C. Marton, F.~Tombari, W.~Wohlkinger, C.~Potthast, B.~Zeisl,
  R.~B. Rusu, S.~Gedikli, and M.~Vincze, ``Point cloud library,'' \emph{IEEE
  Robotics \&amp; Automation Magazine}, vol. 1070, no. 9932/12, 2012.

\bibitem{bishop2006pattern}
C.~M. Bishop, \emph{Pattern recognition and machine learning}.\hskip 1em plus
  0.5em minus 0.4em\relax springer, 2006.

\bibitem{rusu20113d}
R.~B. Rusu and S.~Cousins, ``3d is here: Point cloud library (pcl),'' in
  \emph{Robotics and Automation (ICRA), 2011 IEEE International Conference
  on}.\hskip 1em plus 0.5em minus 0.4em\relax IEEE, 2011, pp. 1--4.

\bibitem{quigley2009ros}
M.~Quigley, K.~Conley, B.~Gerkey, J.~Faust, T.~Foote, J.~Leibs, R.~Wheeler, and
  A.~Y. Ng, ``Ros: an open-source robot operating system,'' in \emph{ICRA
  workshop on open source software}, vol.~3, no. 3.2, 2009, p.~5.

\bibitem{scikit-learn}
F.~Pedregosa, G.~Varoquaux, A.~Gramfort, V.~Michel, B.~Thirion, O.~Grisel,
  M.~Blondel, P.~Prettenhofer, R.~Weiss, V.~Dubourg, J.~Vanderplas, A.~Passos,
  D.~Cournapeau, M.~Brucher, M.~Perrot, and E.~Duchesnay, ``Scikit-learn:
  Machine learning in {P}ython,'' \emph{Journal of Machine Learning Research},
  vol.~12, pp. 2825--2830, 2011.

\bibitem{guyon1997scaling}
I.~Guyon, ``A scaling law for the validation-set training-set size ratio,''
  \emph{AT\&T Bell Laboratories}, pp. 1--11, 1997.

\bibitem{cornelis2008fast}
N.~Cornelis and L.~V. Gool, ``Fast scale invariant feature detection and
  matching on programmable graphics hardware,'' in \emph{IEEE Computer Society
  Conference on Computer Vision and Pattern Recognition Workshops}.\hskip 1em
  plus 0.5em minus 0.4em\relax IEEE, 2008, pp. 1--8.

\end{thebibliography}
